%% file: PaperForReview.tex
\documentclass[10pt,twocolumn,letterpaper]{article}
\pdfoutput=1
\usepackage{cvpr}              

\usepackage{graphicx}
\usepackage{amsmath}
\usepackage{amssymb}
\usepackage{booktabs}
\usepackage[accsupp]{axessibility}

\usepackage[pagebackref,breaklinks,colorlinks]{hyperref}

\usepackage[capitalize]{cleveref}
\crefname{section}{Sec.}{Secs.}
\Crefname{section}{Section}{Sections}
\Crefname{table}{Table}{Tables}
\crefname{table}{Tab.}{Tabs.}

\def\cvprPaperID{10336} 
\def\confName{CVPR}
\def\confYear{2022}
\input{preamble}
\begin{document}

\title{Cross-Domain Correlation Distillation for Unsupervised Domain Adaptation in Nighttime Semantic Segmentation}


\author{Huan Gao\textsuperscript{1}~~~~Jichang Guo\textsuperscript{1}\thanks{Corresponding author}~~~~Guoli Wang\textsuperscript{2}~~~~Qian Zhang\textsuperscript{2}\\
\textsuperscript{1}School of Electrical and Information Engineering, Tianjin University.\\
\textsuperscript{2}Horizon Robotics.\\
{\tt\small \{gh99,~jcguo\}@tju.edu.cn, \{guoli.wang,~qian01.zhang\}@horizon.ai}
}

\maketitle
\begin{abstract}
The performance of nighttime semantic segmentation is restricted by the poor illumination and a lack of pixel-wise annotation, which severely limit its application in autonomous driving. Existing works, \eg, using the twilight as the intermediate target domain to perform the adaptation from daytime to nighttime, may fail to cope with the inherent difference between datasets caused by the camera equipment and the urban style. Faced with these two types of domain shifts, \ie, the illumination and the inherent difference of the datasets, we propose a novel domain adaptation framework via cross-domain correlation distillation, called CCDistill. The invariance of illumination or inherent difference between two images is fully explored so as to make up for the lack of labels for nighttime images. Specifically, we extract the content and style knowledge contained in features, calculate the degree of inherent or illumination difference between two images. The domain adaptation is achieved using the invariance of the same kind of difference. Extensive experiments on Dark Zurich and ACDC demonstrate that CCDistill achieves the state-of-the-art performance for nighttime semantic segmentation. Notably, our method is a one-stage domain adaptation network which can avoid affecting the inference time. Our implementation is available at \url{https://github.com/ghuan99/CCDistill}.
\end{abstract}

\section{Introduction}
\label{sec:intro}

Semantic segmentation as one of the fundamental topics in computer vision, has been widely used in many critical downstream tasks \cite{chen2019learning, geiger2011we}. While a large variety of approaches have been proposed \cite{luc2017predicting, chen2018encoder}, they are predominantly designed to train on daytime images with favorable illumination. However, outdoor applications require satisfactory performance in more challenging scenes, such as nighttime. In this work, we focus on semantic segmentation at nighttime, which is primarily limited by the low exposure of the captured images and the lack of ground truth.

\input{teaser}

To handle this problem, many domain adaptation methods have been proposed to adapt the daytime-trained model to nighttime without requiring ground-truth labels in the nighttime domain. In \cite{wu2021dannet,sakaridis2020map,sakaridis2019guided, romera2019bridging, sun2019see}, they apply an image transfer network to stylize daytime or nighttime images and generate synthetic datasets. However, the style transfer network cannot fully utilize the semantic embedding of the segmentation task and also increases the inference time. Some works \cite{dai2018dark,sakaridis2019guided,sakaridis2020map} utilize the twilight as the intermediate target domain. These methods require additional training data and the training process is complex. Most importantly, all these methods ignore inherent difference between datasets, treating daytime images from different datasets as the same style. Prior work \cite{he2021multi} points out that appearance discrepancy has a significant impact on the effect of adaptation. Ignoring the inherent difference can adversely affect domain adaptation.

Considering the illumination and inherent difference between labeled daytime images and unlabeled nighttime images, we intend to construct an end-to-end multi-source multi-target domain adaptation framework for nighttime semantic segmentation (shown in Fig.~\ref{fig:teaser}). The Dark Zurich \cite{sakaridis2019guided} containing unlabeled daytime ($T_{d}$) and nighttime ($T_{n}$) image pairs and Cityscapes \cite{cordts2016cityscapes} containing labeled daytime images ($S_{d}$) are adopted as our datasets. It can be seen from Fig.~\ref{fig:teaser}, that $T_{d}$ and $T_{n}$ are taken at different times in the close scene, thus there is the huge difference of illumination but highly overlapped semantic information. Although $S_{d}$ and $T_{d}$ are both daytime images, there are obvious differences in the urban style and color tone. We treat the difference in illumination and dataset as the domain shift.

There is a wide literature on knowledge distillation works \cite{yang2021hierarchical, dou2020unpaired, tian2021farewell,wang2021evdistill, hu2021dense, li2021probabilistic} that have explored the cross-modal learning. One of strategies in these methods is to exploit the semantic consistency of images across domains as prior knowledge \cite{tian2021farewell, wang2021evdistill}. However, most of them \cite{li2020towards, dou2020unpaired, tian2021farewell} focus on one teacher and one student. As illustrated in Fig.~\ref{fig:teaser}, we observe that if we can get the $S_{n}$ with content of $S_{d}$ and illumination style of $T_{n}$, the degree of difference in content between $S_{d}$ and $T_{d}$ should be consistent with that between $S_{n}$ and $T_{n}$. Similarly, the degree of difference in illumination or content between $S_{d}$ and $S_{n}$ is consistent with that between $T_{d}$ and $T_{n}$. Therefore, we can leverage the invariance of domain shifts as prior knowledge to implement knowledge distillation in multi-source multi-target domain.

With this insight, we propose a cross-domain correlation distillation approach, which is implemented on the content and style knowledge contained in the feature. The degree of cross-domain difference is obtained by the similarity of two content or style embeddings with only one domain shift, and it can also be regarded as a concrete representation of the domain shift. The cross-domain content correlation is utilized to realize the knowledge distillation from the labeled daytime to the unlabeled nighttime domain, so as to improve the performance of the nighttime semantic segmentation. The premise for the effectiveness of the cross-domain content distillation is that the generated and real nighttime images tend to be as consistent as possible in style. Therefore, we first employ a simple image translation method 
\cite{he2021multi} to align holistic distribution on LAB color space to initially reduce the style discrepancy between day and night. And the cross-domain style distillation can further achieve the style transfer at the semantic-level.

Different from reducing the illumination shift adopted by previous works, it is possible to obtain accurate features of nighttime images by exploiting the consistency of domain shift. We evaluate the performance of CCDistill on Dark Zurich \cite{sakaridis2019guided}, ACDC \cite{sakaridis2021acdc} datasets. Our main contributions are summarized as follows:

\begin{itemize}[leftmargin=*]
\setlength\itemsep{-.3em}
\item For nighttime semantic segmentation, we propose an end-to-end unsupervised domain adaptation framework, CCDistill, which requires neither extra data nor style transfer network, thus it does not affect the inference time of the semantic segmentation network.
\item We propose the cross-domain correlation distillation algorithm, which utilizes the invariance of domain shifts to perform knowledge distillation on content and style embeddings separately. It enables knowledge distillation to be free from the adverse effect caused by the complex domain shifts.
\item Extensive experiments on the Dark Zurich and ACDC datasets verify that our network achieves a new state-of-the-art performance of nighttime semantic segmentation.
\end{itemize}

\input{overview}
\section{Related works}
\label{sec:related}

\paragraph{Domain adaptation}
Domain adaptation can effectively tackle the inconsistent data distribution in different domains. A line of methods utilize the principle of model consistency to reduce the data distribution gap by data augmentation \cite{na2021fixbi}. Chen \etal \cite{chen2021semi} combine source and target domain by the cutmix \cite{yun2019cutmix} and concat. And \cite{ouali2020semi} holds the view that the input level does not follow the cluster assumption, which can be maintained in the embedding space. Therefore, they add different perturbations to the output of the encoder.

However, domain adaptation methods based on style transfer are often more intuitive and integrated \cite{hoffman2018cycada, wu2018dcan}. In \cite{ma2021coarse, he2021multi}, they both convert the source domain image to the LAB color space for style translation. Isobe \etal \cite{isobe2021multi} convert all other domains into the style of the current target domain for further training.

Instead of using data augmentation or style transfer, designing the loss function to constrain the data distribution can also achieve feature alignment \cite{du2021cross, liang2021domain, yue2021prototypical, li2021transferable, hou2021visualizing, hsu2021darcnn}. Wang \etal \cite{wang2021exploring} apply the projection head to map the feature to a 256-d l2-normalized vector, and use the NCE loss on the mapped vector to explore the global semantic relationship. Liu \etal \cite{liu2021source} utilize the KL divergence on the mean and variance stored in the BN layer of the model to make the data distributions similar to each other.

Taking into account the characteristics of the nighttime semantic segmentation, the general domain adaptation methods may fail to cope with the complex domain shift between the daytime and nighttime domains. Therefore, we combine the latter two strategies to construct multi-source and multi-target domains through image-level and semantic-level style transfer, and obtain content embedding by using the JS divergence to constrain data distribution.

\paragraph{Knowledge distillation }
In knowledge distillation (KD), the goal is to transfer additional feedback from the teacher to the student. In early KD methods \cite{hinton2015distilling}, the knowledge transfer is implemented by minimizing the \emph{Kullback-Leibler} (KL) divergence between the predicted distribution of the student and teacher. Recent studies have explored cross-model KD, which transfers high-level knowledge across different modalities \cite{hou2020inter, dai2021general, tian2021farewell,wang2021evdistill, hu2021dense, li2021probabilistic,xu2018pad, wang2019efficient}.

The IntRA-KD \cite{hou2020inter} calculates the mean, variance, and skewness of each category in the feature as the statistics of the current distribution, and uses the cosine similarity of the moment vectors to perform distillation. Similarly, \cite{dai2021general} applies the Euclidean distance to represent the correlation between instances. And \cite{wang2021evdistill, liu2020structured} realize pair-wise distillation by dividing the feature into several nodes and then calculating the similarity between different nodes.

Inspired by the above methods, we explore to make use of the correlation contained in features. The aforementioned methods mostly focus on the situation of a single teacher and a single student. Instead, there are multiple domains in our task, and the inputs of different models include differences in illumination and datasets. Hence, we adopt the distance of embeddings from the two domains with only one kind of domain shift as the high-level representation to transfer knowledge.

\paragraph{Nighttime semantic segmentation}
Previous works on nighttime semantic segmentation apply adversarial models to achieve the style translation from daytime to nighttime \cite{sakaridis2020map, sakaridis2019guided, romera2019bridging, sun2019see, wu2021dannet}. In order to deal with the domain gap, DANNet \cite{wu2021dannet} uses a style translation network to transform different domains as the same style. Besides RGB images, HeatNet \cite{vertens2020heatnet} additionally uses thermal data that is not sensitive to illumination. Many methods adopt twilight as the intermediate domain to gradually reduce the distribution discrepancy \cite{sakaridis2020map, sakaridis2019guided, dai2018dark}. And these methods either require extra data, or need to design additional networks that affect the inference time, and the training process is complicated. Therefore, instead of reducing the illumination shift between labeled daytime and unlabeled nighttime images with a style transfer network, we leverage the domain shift and regard the cross-domain correlation as the concrete representation of the domain shift to realize domain adaptation.

\section{Method}

\subsection{Problem Formulation}
Most existing nighttime semantic segmentation methods mainly consider illumination difference and achieve domain adaptation by reducing the illumination shift between $S_{d}$ and $T_{n}$. While based on our observation, the domain shifts between $S_{d}$ and $T_{n}$ include not only illumination difference but also inherent difference between datasets caused by camera equipment and urban appearance. Regardless of which domain shift is ignored, the effect of domain adaptation will be adversely affected.

In this section, through constructing the multi-source multi-target domain adaptation network, we can select two domains with only one domain shift and calculate the degree of difference between the two domains. Then we propose the cross-domain correlation distillation by using the invariance of cross-domain difference to achieve domain adaptation. Formally, our network involves a source domain $S_{d}$, a synthetic dataset served as another source domain $S_{n}$, and two target domain T, denoted as $T_{d}$ and $T_{n}$, where $D\in\left\{{S_{d},S_{n},T_{d},T_{n}}\right\}$ and these four elements represent Cityscapes (daytime), Cityscapes (synthetic nighttime), Dark Zurich (daytime) and Dark Zurich (nighttime), respectively. Note that only images from $S_{d}$ and $S_{n}$ have the pixel-wise annotation. And $T_{d}$ and $T_{n}$ are taken at different times in the same scene.


The overall architecture of our proposed method is shown in Fig.~\ref{fig:overview}. Our algorithm has four major components:

\begin{itemize}[leftmargin=*]
\setlength\itemsep{-.3em}
\item Semantic Segmentation network. We adopt the RefineNet \cite{lin2017refinenet} as the semantic segmentation network, training two segmentation models $M_{d}$ and $M_{n}$ simultaneously, where $M_{d}$ takes $S_{d}$ and $T_{d}$ as inputs, and $M_{n}$ takes $S_{n}$ and $T_{n}$ as inputs. Our goal is to get the accurate prediction map $\textbf{P}_{T_{n}}$ for $T_{n}$ without using the pixel-level annotation.
\item Project Head. This block is implemented as two 1x1 convolutional layers with ReLU \cite{wang2021exploring}. The intermediate feature $\textbf{F}_{D}$ of $M_{d}$ or $M_{n}$ is input to the project head, and it is mapped to the 256-d l2-normalized vector to extract content embedding for knowledge distillation. Note that this block is only utilized during training, thus the inference time will not be affected.
\item Cross-domain content distillation. We adopt cosine similarity between two content embeddings $\textbf{e}_D$ from different domains to represent the degree of content difference.
\item Cross-domain style distillation. Different from the content knowledge, style embedding $\textbf{G}_D$ is obtained by calculating the Gram matrix \cite{gatys2016image} of the feature $\textbf{F}_D$ itself, and we also use similarity function to measure the cross-domain style difference.
\end{itemize}

\subsection{Cross-domain correlation distillation}
It can be seen from Fig.~\ref{fig:overview}, in addition to the discrepancy of illumination between daytime and nighttime, the daytime images from different datasets also have its particular color tone and urban style. If we can get the $S_{n}$ with content of $S_{d}$ and illumination style of $T_{n}$, the degree of difference in content between $S_{d}$ and $T_{d}$ should be consistent with that between $S_{n}$ and $T_{n}$. Similarly, the degree of difference in illumination or content between $S_{d}$ and $S_{n}$ is consistent with that between $T_{d}$ and $T_{n}$. This invariance of difference in illumination or content can be exploited as prior knowledge to guide the model to extract accurate features for $T_{n}$.

Motivated by the cross-model knowledge distillation \cite{tian2021farewell,wang2021evdistill,dai2021general}, we propose the cross-domain content distillation (CDC) and cross-domain style distillation (CDS). The former conducts the transfer of content knowledge which is essential for the segmentation task, and the latter realizes the style transfer in semantic level.

The following subsections describe in detail how content and style embeddings are extracted and how the degree of difference between the two domains is calculated.

\paragraph{Cross-domain content distillation} The same image always maintains the same semantic content in different styles. Similarly, two images from different datasets should maintain the degree of content difference across styles. The CDC exploits this invariance of content difference to perform semantic knowledge distillation.

Due to the difference in the input of the model $M_{d}$ and $M_{n}$, the feature distribution differs from each other. Here we first utilize the project head to map the features $\textbf{F}_{D}$ into the common embedding space, and get $\textbf{e}_{D}$. Then we further introduce the \emph{Jensen-Shannon} (JS) divergence to constrain the feature distribution. Specifically,
\begin{equation}
\begin{aligned}
L_{JS}=\lambda(JS(\textbf{e}_{S_{d}} || \textbf{e}_{S_{n}})+JS(\textbf{e}_{T_{d}} || \textbf{e}_{T_{n}}))-\\
(JS(\textbf{e}_{S_{d}} || \textbf{e}_{T_{d}})+JS(\textbf{e}_{S_{n}} || \textbf{e}_{T_{n}}))
\end{aligned}
\end{equation}
In order to get the content knowledge contained in the feature, the distribution of embeddings with the same semantic information needs to be close, as the first term in Eq.(1). And at the same time, it is necessary to ensure that the embeddings with different semantic information keep a certain distance, as the second term in Eq. (1). $\lambda$ is the coefficient used to control the effect of reverse JS divergence and it is set to 4.

After getting the content embedding $\textbf{e}_{D}$, we adopt the similarity function to express the degree of content difference between the two domains. In this way, we can get the cross-domain content knowledge, which is formulated as:
\vspace{-0.4cm}
\begin{gather}
Cor_{illu_{k}}=cos(\textbf{e}_{k_{d}},\textbf{e}_{k_{n}}),k\in\left\{ {S,T} \right\}\\
Cor_{in_{r}}=cos(\textbf{e}_{S_{r}},\textbf{e}_{T_{r}}),r\in\left\{{d,n}\right\}
\end{gather}
$Cor_{illu_{k}}$ indicates the correlation of the content within the domain S or the domain T, and $Cor_{in_{r}}$ indicates the inherent correlation between different datasets in the daytime or nighttime scene. $cos(\textbf{x}, \textbf{y}) = \frac{\textbf{x}^T\textbf{y}}{||\textbf{x}||_{2}||\textbf{y}||_{2}}$ is the commonly used cosine similarity. Intuitively, the model which takes the daytime images as input tends to be less difficult to train, and the ground truth in the domain S can also be helpful to extract more superior features. Therefore, $Cor_{illu_{S}}$ is used to guide $Cor_{illu_{T}}$, and $Cor_{in_{d}}$ is used to guide $Cor_{in_{n}}$. We utilize the cross-domain correlation to realize the knowledge transfer from domain S to domain T, from daytime to nighttime, and reduce the disparity of model performance. $Cor_{illu}$ and $Cor_{in}$
represent the patch-level correlation, so they are still effective even if there is the parallax between $T_{d}$ and $T_{n}$. The cross-domain content distillation loss is given as follows:
\begin{small}
\begin{equation}
L_{CDC}=||Cor_{illu_{S}}-Cor_{illu_{T}}||_{2}^{2}+||Cor_{in_{d}}-Cor_{in_{n}}||_{2}^{2}+L_{JS}
\end{equation}
\end{small}
\vspace{-0.2cm}

\noindent The domain shifts that exist between these four domains can be divided into two categories: illumination and inherent difference between different datasets. We select two domains with only one kind of shift each time, construct their correlation graph. For example, $Cor_{in_{d}}$ reflects the similarity between the content of $S_d$ and $T_d$, and there is only the inherent difference between $S_d$ and $T_d$. Similarly, $Cor_{in_{n}}$ reflects the similarity between $S_n$ and $T_n$, and they also have only the inherent difference. $Cor_{in_{d}}$ and $Cor_{in_{n}}$ can be regarded as concrete representations of the inherent difference in the content of daytime and nighttime images between datasets, respectively. Therefore, the process of forcing $Cor_{in_{d}}$ and $Cor_{in_{n}}$ to be equal, as the second term in Eq.~(4), is to utilize the invariance of inherent difference between datasets to achieve the knowledge distillation while avoiding the adverse effects caused by the other kind of domain shift. In a similar way, $Cor_{illu}$ takes advantage of the invariance of content in the same dataset.

\paragraph{Cross-domain style distillation} The premise to implement the CDC is to be able to generate $S_{n}$ with the same illumination style as $T_{n}$. Previous approaches \cite{sakaridis2020map, sakaridis2019guided} generate nighttime images through style translation models, \eg, CycleGAN \cite{zhu2017unpaired}, yet the semantic features in segmentation task are underutilized. 

We first align the mean and variance of $S_{d}$ with $T_{n}$ in the LAB space to get $S_{n}$ \cite{he2021multi, ma2021coarse}. This pre-process can realize the holistic style transformation and decrease the difficulty of model convergence. However, for nighttime images, due to the presence of traffic lights, headlights, etc., there is local overexposure of brightness. If only this holistic style transformation is performed, the generated image will still be quite different from the real nighttime image. As shown in Fig. \ref{fig:overview}, after we perform the moment match in the LAB space, the tone of $S_{n}$ has tended to $T_{n}$ at the holistic level. But for the underexposed or overexposed areas in $T_{n}$, the effect of this style transformation is still not satisfactory. Therefore, we propose the cross-domain style distillation (CDS) to further achieve semantic-level style transfer during the training of segmentation model.

In style transfer \cite{wang2021rethinking, cheng2021style, gatys2016image}, the Gram matrix is used to indicate the self-correlation of features in the channel dimension, which consists of the correlation between the responses of different filters. We adopt the Gram matrix $\textbf{G}_{D} \in R^{C \times C}$ to represent the style of the feature, where $\textbf{G}_{D}$ is the inner product of the vectorised feature maps of $\textbf{F}_{D}$ on channel i and j respectively:
\begin{equation}
\textbf{G}_{D}=\sum_{p}\textbf{F}_{D}^{ip}\textbf{F}_{D}^{jp}, D\in\left\{{S_{d},S_{n},T_{d},T_{n}}\right\}
\end{equation}
where p is the pixel of $\textbf{F}_{D}$. After obtaining the style knowledge of the feature $\textbf{F}_{D}$ itself, the principle of our style transfer is similar to that of the CDC. We also build the cross-domain style graph, and this can be formulated as:
\begin{gather}
Cor_{G_{k}}=cos(\textbf{G}_{k_{d}},\textbf{G}_{k_{n}}),k\in\left\{{S,T}\right\}\\
L_{CDS}=||Cor_{G_{S}}-Cor_{G_{T}}||_{2}^{2}
\end{gather}
$Cor_{G_{k}}$ reflects the degree of illumination difference in the source or target domain. The alignment of style difference between domain S and T achieves the semantic-level style transfer, as defined in Eq. (7).

It is worth noting that only the style correlation in the domain $S$ or the domain $T$ is used here. The main reason is that although there is inherent style shift between $S_{d}$ and $T_{d}$, this is less noticeable than the illumination difference between the daytime and nighttime, thus the correlation of the Gram matrices between them is not strong. The CDS mainly utilizes the invariance of illumination difference between daytime and nighttime images in the same dataset to perform style transfer, so that the $S_{n}$ gradually approaches the illumination style of $T_{n}$. And this choice can also exclude the adverse effect caused by the inherent difference.

\subsection{Objective functions}
In summary, the total loss of our method is written as follows:
\begin{small}
\begin{equation}
L=L_{seg_{n}}+L_{seg_{d}}+L_{pseudo}+\lambda_{1}L_{CDC}+\lambda_{2}L_{CDS}
\end{equation}
\end{small}
\vspace{-0.2cm}

\noindent where $L_{seg_{n}}$ is the weighted cross-entropy loss between the prediction map $\textbf{P}_{S_n}$ and the corresponding ground truth, and $L_{seg_{d}}$ is in the same way. $L_{pseudo}$ is the static loss \cite{wu2021dannet}, which uses the predictions of static object categories for the daytime images $T_d$ as the pseudo labels to provide pixel-level supervision on $T_n$. The $\lambda_{1}$, $\lambda_{2}$ are hyper-parameters that balance the influence of distillation losses on the main task, which are set to 2 and 1, respectively.

\input{sota}
\input{val}
\input{comparisons}

\section{Experiments}
\subsection{Datasets}
The following datasets are used for model training and performance evaluation:

\noindent\textbf{Cityscapes} \cite{cordts2016cityscapes} is an autonomous driving dataset captured from street scenes in 50 cities, with pixel-wise annotations of 19 semantic categories. It contains 2,975 images for training, 500 images for validation and 1,525 images for testing. All images are at a fixed resolution of 2,048×1,024. In this paper, we adopt the Cityscapes training set in the training of our network.

\noindent\textbf{Dark Zurich} \cite{sakaridis2019guided} is captured in Zurich, with 3,041 daytime, 2,920 twilight and 2,416 nighttime images for training, which are all unlabeled with a resolution of 1,920x1,080. Each nighttime image has a corresponding daytime image as auxiliary, which constitutes a data pair that can be used for the knowledge distillation in our proposed network. Thus we use the 2,416 night-day image pairs in our training process. The Dark Zurich also contains 201 manually annotated nighttime images, of which 151 (Dark Zurich-test) are used for testing and 50 (Dark Zurich-val) are used for validation. Note that the evaluation of Dark Zurich-test only serves as an online benchmark, and its ground truth is not publicly available.

\noindent\textbf{ACDC} \cite{sakaridis2021acdc} consists of 4006 images including four common adverse conditions: fog, rain, nighttime and snow. The images under nighttime scenes have pixel-wise annotations, and are further divided into 400 training, 106 validation and 500 test images. This dataset and the Dark Zurich are both proposed by Sakaridis \etal \cite{sakaridis2019guided}, thus it shares the similar style and appearance with the Dark Zurich. So we adopt the ACDC-night-val to further evaluate the effect of our network for domain adaptation.

\subsection{Implementation details}
We implement the proposed network using PyTorch on a single Titan RTX GPU. We adopt the RefineNet \cite{lin2017refinenet} as our semantic segmentation model, which is pre-trained on the Cityscapes dataset with the ResNet-101 \cite{he2016deep} as backbone. Both our models are trained by the Stochastic Gradient Descent (SGD) optimizer with a momentum of 0.9 and a weight decay of $5\times10^{-4}$, and the initial learning rate is set as $2.5\times10^{-4}$. Then the learning rate is decreased with the poly policy with a power of 0.9. The batch size is set to 2. The total number of training iterations is 50k. Following \cite{wu2021dannet}, we apply random cropping with a crop size of 512 for Cityscapes dataset, and with a crop size of 960 for Dark Zurich which is then resized to 512. At the inference time, there is no any change introduced to the final model $M_{n}$.

\input{acdc}

\subsection{Comparison with state-of-the-art methods}
\vspace{-0.2cm}
\paragraph{Comparison on Dark Zurich} We compare our proposed method with some existing state-of-the-art methods, including DMAda \cite{dai2018dark}, GCMA \cite{sakaridis2019guided}, MGCDA \cite{sakaridis2020map}, DANNet \cite{wu2021dannet}, and several other domain adaptation approaches \cite{tsai2018learning, vu2019advent, li2019bidirectional, toldo2021unsupervised} on Dark Zurich-test. The MGCDA, GCMA, DMAda and DANNet adopt the RefineNet \cite{lin2017refinenet} as the baseline, while other methods use the Deeplab-v2 \cite{chen2017deeplab}. To ensure a fair comparison, we perform our method on the RefineNet. Note that both the baselines use the ResNet-101 \cite{he2016deep} as backbone. \Table{sota} shows the quantitative comparison with other methods on Dark Zurich-test. The mIoU is calculated by the average of the intersection-over-union (IoU) among all 19 categories.

Our method surpasses the existing methods with around 3.2$\%$ increase on mIoU. In particular, CCDistill is a one-stage adaptation framework with requiring no additional network in the inference. We also observe that our approach has comparable effects in all large-scale categories such as terrain, sidewalk and road, which proves that our method achieves the style transfer from daytime to nighttime and thus realizes the cross-domain knowledge distillation. Moreover, CCDistill significantly improves the performance of categories with relatively few occurrences, such as train and motorcycle. This also indicates that our method does transfer the semantic-level correlation knowledge. The qualitative results on Dark Zurich-val, as shown in Fig. \ref{fig:val}, can also verify this observation.

\paragraph{Comparison on ACDC} In order to verify the effectiveness of the proposed model on nighttime semantic segmentation, we further conduct comparative experiments on the ACDC-night-val, and the results are shown in \Table{comparisons}. The ACDC-night has a similar nighttime style with the Dark Zurich-night, so it is reasonable that the CCDistill achieves the best performance on ACDC-night-val, and a 0.7$\%$ improvement of mIoU is gained. The visualization comparison on ACDC-night-val is shown in Fig. \ref{fig:acdc}. 

\subsection{Generalization test}
Same as the daytime images, there are also domain shifts between nighttime images from different datasets. In order to verify the generalization of our proposed method, we also compare with other methods on the BDD100K-night. The BDD100K-night contains 87 images with the resolution of 1,280×720, which is manually selected by \cite{sakaridis2020map} from the 345 nighttime images of BDD100K \cite{yu2018bdd100k}. The appearance and lighting tone between Dark Zurich and BDD100K-night are quite different. As shown in \Table{comparisons}, even though the target domain of our proposed method is Dark Zurich and it is the domain shift between Cityscapes and Dark Zurich that we utilize, we still get the comparable performance on the BDD100K-night.

\input{ablations}

\subsection{Ablation study}
In this section, extensive experiments on several model variants are conducted to verify the effectiveness of each proposed component. We measure the performance of each ablated version by evaluating it on the Dark Zurich-test. Results are summarized in \Table{ablations}.

The content correlation across domains is the core of the knowledge distillation in our method. We set up five ablated version to prove the effect of the proposed CDC. First, removing the CDC and relying only on CDS for knowledge distillation lead to a drop of 2.2$\%$ mIoU. We further assess the role of each component in the CDC. Training without the project head astonishingly deteriorates the mIoU by 9.3$\%$, which verifies that the difference in feature distribution caused by the domain gap in the task will seriously affect the effectiveness of knowledge distillation. Features from different domains need to be mapped to the common embedding space to approximate the distribution range, and then the effective correlation knowledge can be extracted. On the basis of the project head, $L_{JS}$ is conducive to further obtaining the content embedding.  Experiment shows that disabling the $L_{JS}$ causes a 2.9$\%$ mIoU decrease. Subsequently, $Cor_{illu}$ and $Cor_{in}$ will be used in the CDC to realize the distillation of the 
content correlation across domains, which contribute 1.8$\%$ and 1.6$\%$ mIoU respectively. Note that after removing the project head or $L_{JS}$ in the CDC, the performance is worse than disabling the CDC completely. This proves from the side that CDS has realized the satisfactory style transfer which is beneficial for the nighttime semantic segmentation. More importantly, it reflects that knowledge distillation in the domain adaptation is very sensitive that the failure to extract the appropriate embedding will be detrimental to the model. In summary, these model variants verify that for domain adaptation with large domain shifts, the adequate and effective use of correlation knowledge within a similar range of data distribution can greatly improve the performance of the model.

The same illumination style between the $S_{n}$ and $T_{n}$ is the prerequisite for the cross-domain content distillation. We disable the CDS resulting in a drop of 3.5$\%$ mIoU, which is in line with expectations. The semantic-level style alignment implemented by CDS can generate nighttime images aiming at the nighttime semantic segmentation, and obtain the synthetic domain that satisfies our hypothetical domain shift. The LAB-based translation advances the performance about 4$\%$, which reflects that the holistic style alignment can reduce the difficulty of subsequent semantic-level transfer. This also proves that it is not appropriate to employ distillation loss directly when the domain shift is large.

After removing CDC and CDS, the model achieves 44.8$\%$ mIoU. The CDS brings a gain of 0.5$\%$ mIoU, while only adding CDC, mIoU has dropped by 0.8$\%$. This further illustrates the importance of CDS to achieve semantic-level style transfer, and on the basis of CDS, CDC can further achieve a huge improvement.

\section{Conclusions}
In this paper, we propose an unsupervised domain adaptation framework via the invariance of cross-domain difference for nighttime semantic segmentation. We validate the effectiveness of properly handling these two kind of domain shifts, \ie illumination and inherent difference. The proposed cross-domain content and style distillation, by extracting the content and style knowledge contained in the features, utilize the invariance of inherent and illumination difference across domains, and realize knowledge distillation and sementic-level style transfer simultaneously. Experiment results verify the effectiveness of our proposed method. Since the distillation is based on the domain shift between source and target domain, it cannot always be effective enough for all nighttime style, which will be further explored in our future work.

\noindent\textbf{Acknowledgement}
This work is supported by the National Nature Science Foundation of China (No.62171315).

\newpage
{\small
\bibliographystyle{ieee_fullname}
\bibliography{PaperForReview}
}

\end{document}

%% file: preamble.tex
\usepackage{overpic}
\usepackage{enumitem} 
\usepackage{overpic} 
\usepackage{color}

\definecolor{turquoise}{cmyk}{0.65,0,0.1,0.3}
\definecolor{purple}{rgb}{0.65,0,0.65}
\definecolor{dark_green}{rgb}{0, 0.5, 0}
\definecolor{orange}{rgb}{0.8, 0.6, 0.2}
\definecolor{red}{rgb}{0.8, 0.2, 0.2}
\definecolor{darkred}{rgb}{0.6, 0.1, 0.05}
\definecolor{blueish}{rgb}{0.0, 0.3, .6}
\definecolor{light_gray}{rgb}{0.7, 0.7, .7}
\definecolor{pink}{rgb}{1, 0, 1}
\definecolor{greyblue}{rgb}{0.25, 0.25, 1}

\newcommand{\Table}[1]{Table~\ref{tab:#1}}

\usepackage{blindtext}

\renewcommand{\paragraph}[1]{\vspace{1em}\noindent\textbf{#1}.}

%% file: teaser.tex
\begin{figure}[t]
\begin{center}
\includegraphics[width=\linewidth]{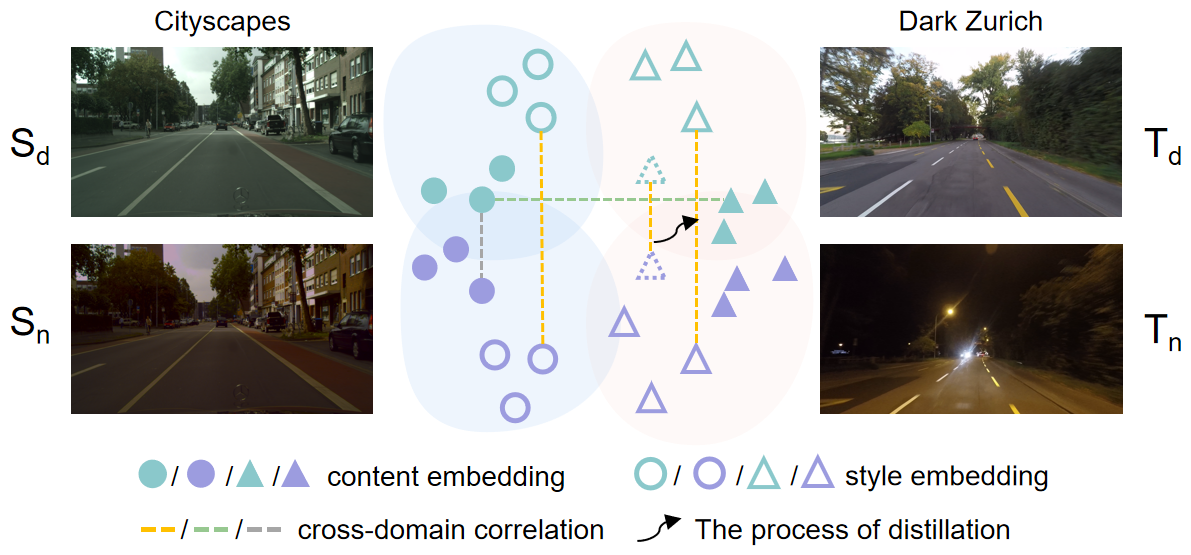}
\end{center}
\setlength{\abovecaptionskip}{1pt}
\caption{
Every two embeddings connected by the dotted line come from two domains, and they only have one difference in illuminance (\ie, each column) or dataset (\ie, each row). The cross-domain correlation reflects the similarity of the two domains, and can also be considered as a concrete representation of the domain shift. Here we only illustrate the cross-domain style distillation. Our main idea is to make the different cross-domain correlations under the same domain shift consistent. 
}
\setlength{\belowcaptionskip}{1pt}
\vspace{-0.2cm}
\label{fig:teaser}
\end{figure}

%% file: overview.tex
\begin{figure*}
\begin{center}
\includegraphics[width=\linewidth]{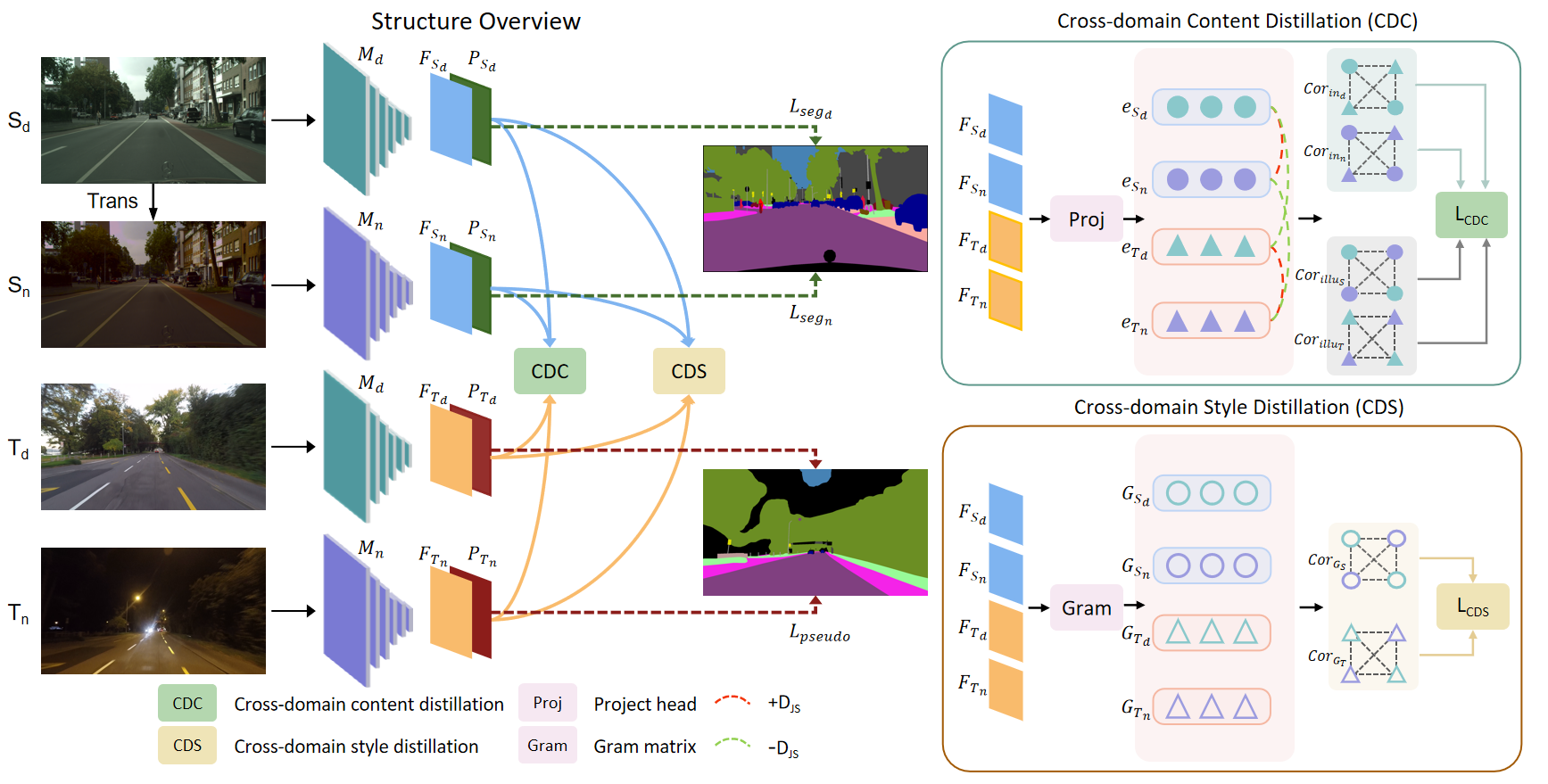}
\hfill
\end{center}
\setlength{\abovecaptionskip}{1pt}
\caption{
Framework. 1) The overview of our proposed framework is shown on the left. The architecture consists of two semantic segmentation models $M_{d}$ and $M_{n}$. The colored solid arrows represent the data flow of the middle layer features  $\textbf{F}_D$ from different domains, and the colored dash arrows represent the supervision to the outputs  $\textbf{P}_D$. 2) The specific distillation is shown on the right. The  $\textbf{e}_D$ and  $\textbf{G}_D$ represent content and style embedding, respectively. For  $\textbf{F}_D$,  $\textbf{P}_D$,  $\textbf{e}_D$ and  $\textbf{G}_D$, the subscript indicates which domain they are obtained from.
}
\setlength{\belowcaptionskip}{1pt}
\label{fig:overview}
\end{figure*}

%% file: sota.tex
\begin{table*}
\centering
\large
\setlength{\abovecaptionskip}{1pt}
\caption{
Comparison with the state-of-the-art approaches and baseline models on the Dark Zurich-test set.
} 
\setlength{\belowcaptionskip}{1pt}
\label{tab:sota}
\setlength{\tabcolsep}{1mm}{
\resizebox{\textwidth}{!}{ 
\begin{tabular}{@{}lcccccccccccccccccccc@{}}
\specialrule{0em}{1.2pt}{1.2pt}
\toprule
Method & \rotatebox{90}{road} & \rotatebox{90}{sidewalk} & \rotatebox{90}{building} & \rotatebox{90}{wall} & \rotatebox{90}{fence} & \rotatebox{90}{pole} & \rotatebox{90}{traffic light} & \rotatebox{90}{traffic sign} & \rotatebox{90}{vegetation} & \rotatebox{90}{terrain} & \rotatebox{90}{sky} & \rotatebox{90}{person} & \rotatebox{90}{rider} & \rotatebox{90}{car} & \rotatebox{90}{truck} & \rotatebox{90}{bus} & \rotatebox{90}{train} & \rotatebox{90}{motorcycle} & \rotatebox{90}{bicycle} & mIoU\\
\midrule
DeepLab-v2-Cityscapes \cite{chen2017deeplab} & 79.0 & 21.8 & 53.0 & 13.3 & 11.2 & 22.5 & 20.2 & 22.1 & 43.5 & 10.4 & 18.0 & 37.4 & 33.8 & 64.1 & 6.4 & 0.0 & 52.3 & 30.4 & 7.4 & 28.8 \\
RefineNet-Cityscapes \cite{lin2017refinenet} & 68.8 & 23.2 & 46.8 & 20.8 & 12.6 & 29.8 & 30.4 & 26.9 & 43.1 & 14.3 & 0.3 & 36.9 & 49.7 & 63.6 & 6.8 & \underline{0.2} & 24.0 & 33.6 & 9.3 & 28.5 \\
\midrule
AdaptSegNet-Cityscapes→DZ-night \cite{tsai2018learning} & 86.1 & 44.2 & 55.1 & 22.2 & 4.8 & 21.1 & 5.6 & 16.7 & 37.2 & 8.4 & 1.2 & 35.9 & 26.7 & 68.2 & 45.1 & 0.0 & 50.1 & 33.9 & 15.6 & 30.4 \\
ADVENT-Cityscapes→DZ-night \cite{vu2019advent} & 85.8 & 37.9 & 55.5 & 27.7 & 14.5 & 23.1 & 14.0 & 21.1 & 32.1 & 8.7 & 2.0 & 39.9 & 16.6 & 64.0 & 13.8 & 0.0 & 58.8 & 28.5 & 20.7 & 29.7 \\
BDL-Cityscapes→DZ-night \cite{li2019bidirectional} & 85.3 & 41.1 & 61.9 & 32.7 & 17.4 & 20.6 & 11.4 & 21.3 & 29.4 & 8.9 & 1.1 & 37.4 & 22.1 & 63.2 & 28.2 & 0.0 & 47.7 & 39.4 & 15.7 & 30.8 \\
UDAclustering-Cityscapes→DZ-night \cite{toldo2021unsupervised} & 85.5 & 40.9 & 59.2 & 31.2 & 19.5 & 24.0 & 29.9 & 29.4 & 30.6 & 11.2 & 18.4 & 39.1 & 49.7 & 61.5 & 34.9 & 0.0 & 25.8 & 23.2 & 19.0 & 33.3\\
DMAda \cite{dai2018dark} & 75.5 & 29.1 & 48.6 & 21.3 & 14.3 & 34.3 & 36.8 & 29.9 & 49.4 & 13.8 & 0.4 & 43.3 & \underline{50.2} & 69.4 & 18.4 & 0.0 & 27.6 & 34.9 & 11.9 & 32.1 \\
GCMA \cite{sakaridis2019guided} & 81.7 & 46.9 & 58.8 & 22.0 & 20.0 & \underline{41.2} & \textbf{40.5} & \textbf{41.6} & 64.8 & \underline{31.0} & 32.1 & \textbf{53.5} & 47.5 & \textbf{75.5} & 39.2 & 0.0 & 49.6 & 30.7 & 21.0 & 42.0 \\
MGCDA \cite{sakaridis2020map} & 80.3 & 49.3 & 66.2 & 7.8 & 11.0 & \textbf{41.4} & \underline{38.9} & \underline{39.0} & 64.1 & 18.0 & 55.8 & \underline{52.1} & \textbf{53.5} & \underline{74.7} & \textbf{66.0} & 0.0 & 37.5 & 29.1 & \underline{22.7} & 42.5 \\
DANNet(RefineNet) \cite{wu2021dannet} & \textbf{90.0} & \underline{54.0} & \textbf{74.8} & \textbf{41.0} & \underline{21.1} & 25.0 & 26.8 & 30.2 & \textbf{72.0} & 26.2 & \textbf{84.0} & 47.0 & 33.9 & 68.2 & 19.0 & \textbf{0.3} & \underline{66.4} & \underline{38.3} & \textbf{23.6} & \underline{44.3} \\
\midrule
\textbf{Ours} & \underline{89.6} & \textbf{58.1} & \underline{70.6} & \underline{36.6} & \textbf{22.5} & 33.0 & 27.0 & 30.5 & \underline{68.3} & \textbf{33.0} & \underline{80.9} & 42.3 & 40.1 & 69.4 & \underline{58.1} & 0.1 & \textbf{72.6} & \textbf{47.7} & 21.3 & \textbf{47.5}\\
\bottomrule

\end{tabular}
} 
}
\end{table*}

%% file: val.tex
\begin{figure*}
\begin{center}
\includegraphics[width=\linewidth]{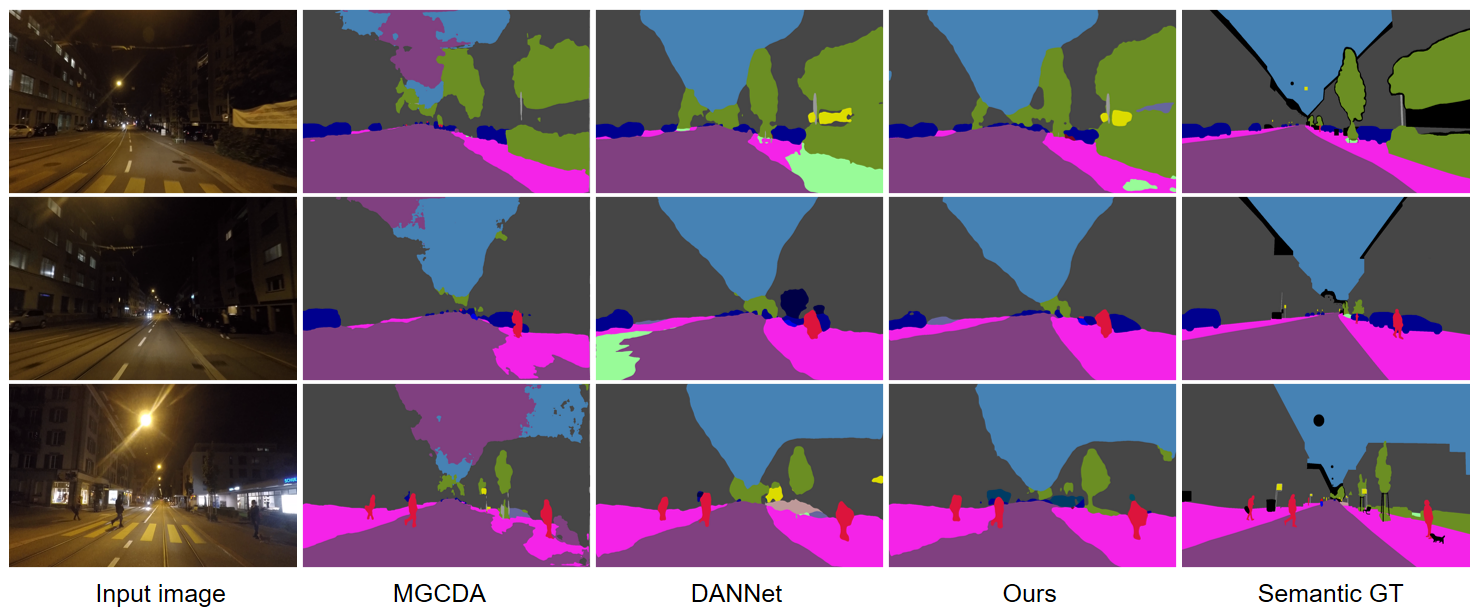}
\hfill
\end{center}
\setlength{\abovecaptionskip}{1pt}
\caption{
The qualitative comparison between our approach and some existing state-of-the-art methods on the Dark Zurich-val set.
}
\setlength{\belowcaptionskip}{1pt}
\vspace{-0.2cm}
\label{fig:val}
\end{figure*}

%% file: comparisons.tex
\begin{table}
\centering
\setlength{\tabcolsep}{0.5mm}{ 
\setlength{\abovecaptionskip}{1pt}
\caption{
Comparison with the state-of-the-art methods and baseline models on the ACDC-night-val set (mIoU1) and the BDD100K-night set (mIoU2).
} 
\setlength{\belowcaptionskip}{1pt}
\label{tab:comparisons}
\begin{tabular}{@{}lccc@{}}
\toprule
Method & mIoU1 & mIoU2 \\
\midrule
DeepLab-v2-Cityscapes \cite{chen2017deeplab} & 16.3 & 17.3 \\
RefineNet-Cityscapes \cite{lin2017refinenet} & 20.3 & 20.4 \\
\midrule
AdaptSegNet-Cityscapes→DZ-night \cite{tsai2018learning} & 23.8 & 22.0 \\
ADVENT-Cityscapes→DZ-night \cite{vu2019advent} & 26.2 & 22.6 \\
BDL-Cityscapes→DZ-night \cite{li2019bidirectional} & 23.9 &  22.8 \\
UDAclustering-Cityscapes→DZ-night \cite{toldo2021unsupervised} & 24.5 & 20.0\\
DMAda \cite{dai2018dark} & - &  28.3 \\
GCMA \cite{sakaridis2019guided} & - & \underline{33.2} \\
MGCDA \cite{sakaridis2020map} & 29.0 & \textbf{34.9} \\
DANNet(RefineNet) \cite{wu2021dannet} & \underline{37.0} &  30.3 \\
\midrule
\textbf{Ours} & \textbf{37.7}  & 33.0\\
\bottomrule
\end{tabular}
} 
\end{table}

%% file: acdc.tex
\begin{figure*}
\begin{center}
\includegraphics[width=\linewidth]{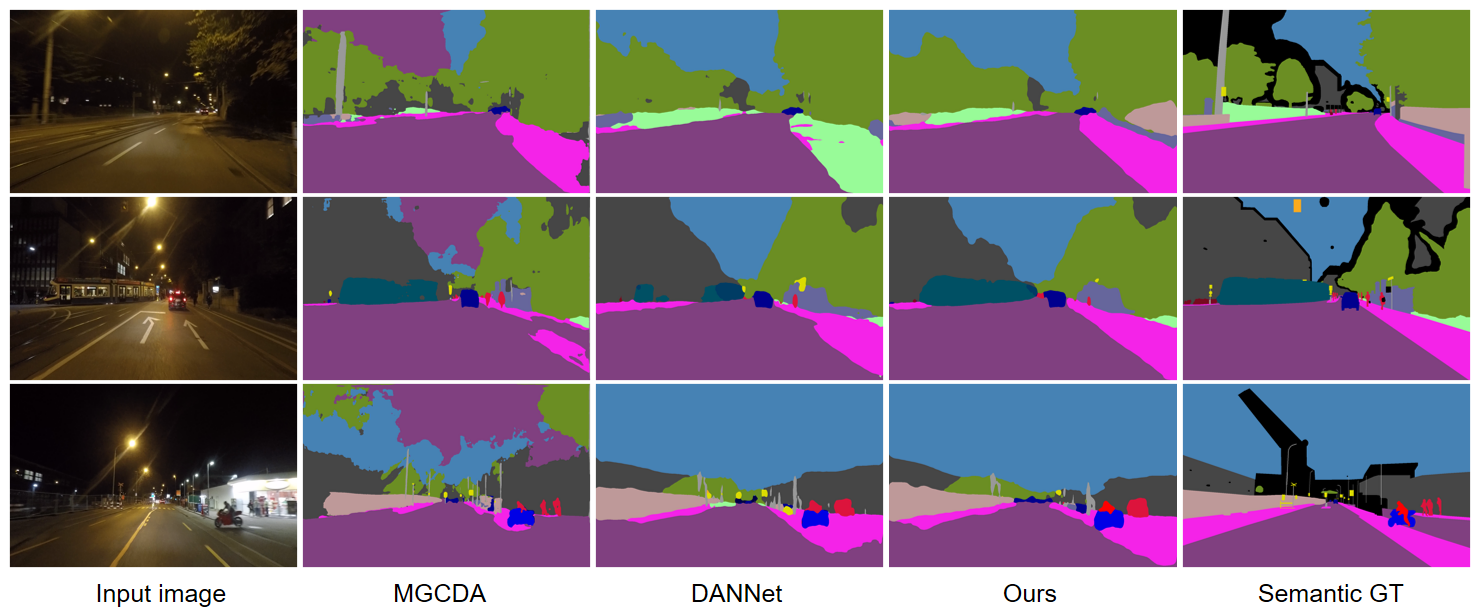}
\hfill
\end{center}
\setlength{\abovecaptionskip}{1pt}
\caption{
The qualitative comparison between our approach and some existing state-of-the-art methods on the ACDC-night-val set.
}
\setlength{\belowcaptionskip}{1pt}
\vspace{-0.2cm}
\label{fig:acdc}
\end{figure*}

%% file: ablations.tex
\begin{table}
\centering
\setlength{\tabcolsep}{12mm}{ 
\setlength{\abovecaptionskip}{1pt}
\caption{
Ablation studies of our proposed method on Dark Zurich-test set.
} 
\setlength{\belowcaptionskip}{1pt}
\label{tab:ablations}
\begin{tabular}{@{}lc@{}}
\toprule
Method & mIoU \\
\midrule
RefineNet &  28.5\\
\midrule
w/o CDC &  45.3\\
w/o project head in CDC &  38.2\\
w/o $L_{JS}$ in CDC &  44.6\\
w/o illuminance correlation in CDC &  45.7\\
w/o inherent correlation in CDC &  45.9\\
\midrule
w/o CDS & 44.0\\
w/o LAB-based Trans &  43.5\\
\midrule
w/o CDC and CDS & 44.8 \\
\midrule
\textbf{Ours} & 47.5\\
\bottomrule
\end{tabular}
} 
\end{table}